\documentclass{article}
\usepackage{ijcai23}

\usepackage{times}
\usepackage{soul}
\usepackage{url}
\usepackage[hidelinks]{hyperref}
\usepackage[utf8]{inputenc}
\usepackage[small]{caption}
\usepackage{graphicx}
\usepackage{amsmath}
\usepackage{amsthm}
\usepackage{booktabs}
\usepackage{algorithm}
\usepackage{algorithmic}
\usepackage[switch]{lineno}
\usepackage{amsfonts}
\usepackage{multirow}
\usepackage{subfigure}
\usepackage{bbding}
\usepackage{xcolor}

\newtheorem{definition}{Definition}
\begin{document}
	
	\title{FedNoRo: Towards Noise-Robust Federated Learning by Addressing Class Imbalance and Label Noise Heterogeneity}
	\author{
		Nannan Wu$^1$
		\and
		Li Yu$^1$\and
		Xuefeng Jiang$^2$\and
		Kwang-Ting Cheng$^3$\And
		Zengqiang Yan$^1$\thanks{Corresponding author}
		\affiliations
		$^1$School of Electronic Information and Communications, Huazhong University of Science and Technology\\
		$^2$Institute of Computing Technology, Chinese Academy of Sciences\\
		$^3$School of Engineering, Hong Kong University of Science and Technology\\
		\emails
		\{wnn2000, hustlyu\}@hust.edu.cn,
		jiangxuefeng21b@ict.ac.cn,
		timcheng@ust.hk,
		z\_yan@hust.edu.cn
	}
	\maketitle
	
	\begin{abstract}
		Federated noisy label learning (FNLL) is emerging as a promising tool for privacy-preserving multi-source decentralized learning. Existing research, relying on the assumption of class-balanced global data, might be incapable to model complicated label noise, especially in medical scenarios. In this paper, we first formulate a new and more realistic federated label noise problem where global data is class-imbalanced and label noise is heterogeneous, and then propose a two-stage framework named FedNoRo\footnote{Code is available at \url{https://github.com/wnn2000/FedNoRo}.} for noise-robust federated learning. Specifically, in the first stage of FedNoRo, per-class loss indicators followed by Gaussian Mixture Model are deployed for noisy client identification. In the second stage, knowledge distillation and a distance-aware aggregation function are jointly adopted for noise-robust federated model updating. Experimental results on the widely-used ICH and ISIC2019 datasets demonstrate the superiority of FedNoRo against the state-of-the-art FNLL methods for addressing class imbalance and label noise heterogeneity in real-world federated learning scenarios. 
	\end{abstract}
	
	\section{Introduction} \label{section1}
	
	Federated Learning (FL), allowing individual clients to train a deep learning model collaboratively without data sharing, has been widely studied in privacy-conscious occasions, e.g. medical \cite{kaissis2020secure} and financial \cite{zheng2021federated} applications.
	Most existing federated learning frameworks are based on the paradigm of fully supervised learning \cite{mcmahan2017communication,FedProx}, which implicitly assumes that each participant's data is labeled correctly. 
	However, building a completely clean dataset with high-quality annotation is costly in realistic medical scenarios, as labeling medical data is time-consuming and labor-intensive requiring expertise. Consequently, it would unavoidably introduce noisy labels when hiring non-professionals to label or using automatic labeling techniques \cite{irvin2019chexpert}. Therefore, it is more common that the labels of some clients are clean while others are not in real-world FL scenarios.
	
	Due to the existence of noisy clients, developing a noise-robust FL framework is of great importance, where accurately identifying the noisy clients is the first and the most crucial step. Existing methods for noisy client detection propose to calculate an average indicator (\textit{e.g.} loss) over all samples of each client as its feature and filter out the clients with abnormal features as noisy clients, which assumes clean clients' features are independent and identically distributed (IID) while noisy clients' are outliers following the small-loss trick \cite{han2018co}. Specifically, Xu \emph{et al.} \shortcite{xu2022fedcorr} calculated the average LID value \cite{houle2013dimensionality} of each client and identified the clients with larger average LID values as noisy clients. Similarly, Wang \emph{et al.} \shortcite{wang2022fednoil} replaced the LID value with the confidence score and identified noisy clients with smaller confidence scores. Unfortunately, these indicators can be less effective to deal with noisy clients in real-world FL scenarios, due to the following observations:
	\begin{enumerate}
		\item Data is highly class-imbalanced from the global perspective \cite{wang2021addressing,shang2022federated}. Under class imbalance, using a global indicator for all classes is highly sensitive to clients' label distributions which may vary dramatically.
		
		\item Data is heterogeneous across clients \cite{FedProx}. As each client collects its own data independently under FL, there may exist severe data variations, affecting the calculation of indicators across clients.
		
		\item Label noise is heterogeneous across clients where the heterogeneity of noise varies in both strength and pattern. The former represents different noise rates across clients, and the latter indicates various forms of label noise related to clients' local data distributions.
	\end{enumerate}
	Suffering from class imbalance, the federated model in FL can bias to the global majority classes \cite{kang2019decoupling,zhou2020bbn,cao2019learning,menon2020long}, resulting in large indicator variations across classes and making noisy clients' indicators less distinguishable. 
	For instance, in clinical scenarios, given one cancer-specialized hospital \textit{A} and one general hospital \textit{B}, \textit{A} would enroll more malignant cases and \textit{B} will enroll more healthy cases. Therefore, \textit{A} is more likely to produce an abnormal client-wise feature (\textit{e.g.}, large loss values similar to noisy clients) due to class imbalance (\textit{i.e.}, healthy $\gg$ malignant). 
	In addition to data heterogeneity, the client-wise feature can also be affected by heterogeneous label noise, making noisy client detection harder. 
	For instance, in clinical scenarios, given two hospitals \textit{C} and \textit{D} with different patient diversity (\textit{i.e.}, various label distributions), one benign case is more likely to be wrongly diagnosed as healthy by \textit{C} and malignant by \textit{D}. Though both labels are wrong, the loss values (\textit{i.e.}, indicators for noisy client detection) produced by \textit{C} would be much smaller than \textit{D}, due to class imbalance (\textit{i.e.}, healthy $\gg$ malignant).
	In summary, class imbalance would divert the client-wise feature from being only related to whether a client is noisy to being related to both its local data distribution and label noise pattern, making the above identification methods based on global indicator struggle in realistic FL scenarios.
	
	To address this, one straightforward way is to combine class-balancing \cite{kang2019decoupling,zhou2020bbn,cao2019learning,menon2020long} with noisy label learning. However, those methods can only mitigate the bias in class prior probability while the bias in learning difficulty of class-specific features \cite{yi2022identifying} is unsolved. Due to relatively limited training data, the class-specific features of minority classes would be more difficult to learn compared to those of the majority classes, resulting in imbalance in the feature space. When adopting feature learning \cite{karthik2021learning,yi2022identifying} to alleviate the bias in learning difficulty, it can be constrained by each client's limited local data and computing resources. Till now, how to address class imbalance in federated noisy label learning (FNLL) is under-explored.
	
	In this paper, we first formulate a new FNLL problem to model more realistic FL scenarios under class imbalance, and then propose a two-stage framework FedNoRo for noise-robust learning. Specifically, in the first stage, instead of using a global indicator, we propose to identify noisy clients according to the client-wise per-class average loss values calculated by a warm-up model trained by FedAvg \cite{mcmahan2017communication}. As each class is considered independently, this detection method will be less affected by class imbalance and heterogeneity, leading to better detection performance. In the second stage, different learning strategies are employed for clean and noisy clients respectively, where cross-entropy loss is adopted for training on clean clients and knowledge distillation (KD) \cite{hinton2015distilling} is employed to minimize the negative influence of noisy labels on noisy clients. In addition to client-level training, a distance-aware aggregation function is proposed to better balance the importance of clean and noisy clients for global model updating in the server. In the local training phase of both the two stages, logit adjustment (LA) \cite{menon2020long} is imposed to fight against data heterogeneity and class imbalance. The main contributions are summarized as follows:
	
	\begin{itemize}
		\item A new FNLL problem where both class-imbalanced global data and heterogeneous label noise are considered to model real FL scenarios.
		
		\item A new label noise generation approach for multi-source data, where the synthetic label noise is heterogeneous and instance-dependent.
		
		\item A two-stage FL framework, named FedNoRo, to address both class imbalance and label noise heterogeneity. In FedNoRo, noisy clients are identified based on abnormal per-class loss values, and noise-robust training strategies are used for effective federated model updating.
		
		\item Superior performance against the state-of-the-art FNLL approaches on real-world multi-source medical datasets.
		
	\end{itemize}
	
	\section{Related Work}
	
	\subsection{Federated Learning}
	
	Federated learning (FL) has drawn great attention for privacy-preserving applications \cite{mcmahan2017communication}. Existing studies mainly focus on the challenges of FL applications, including data heterogeneity \cite{FedProx,karimireddy2020scaffold,li2021model,li2021fedbn,yan2020variation,zhu2021data,li2021fedrs}, data quality \cite{liu2021federated,liang2022rscfed,jiang2022dynamic}, privacy protection \cite{agarwal2018cpsgd} and communication efficiency \cite{konevcny2016federated,sattler2019robust}.
	
	\subsection{Noisy Label Learning}
	
	Noisy label learning is of great importance as it is difficult to collect a large number of high-quality clean labels in real-world applications. Existing studies can be roughly divided into two types: selection-based \cite{han2018co,yu2019does,li2020dividemix,wei2022self,zhao2022centrality,yi2022identifying} and selection-free \cite{wang2019symmetric,zhang2017mixup,wei2021smooth,lukasik2020does,lukov2022teaching}. The former assumes that clean and noisy samples behave differently in a certain indicator (\textit{e.g.} loss value) which can be used for identification. The latter focuses on designing special loss functions or regularizations that are robust to noisy labels. 
	
	\subsection{Federated Noisy Label Learning}
	
	Federated noisy label learning (FNLL) is an emerging research topic. Chen \emph{et al.} \shortcite{chen2020focus} and Yang \emph{et al.} \shortcite{yang2021client} leveraged clean datasets to assist FNLL, which may be infeasible in practice. RoFL \cite{yang2022robust} updated local models with shared class-wise centroids to avoid inconsistent decision boundaries caused by noisy labels. FedLSR \cite{jiang2022towards} introduced regularization in the local training phase to alleviate the effect of noisy labels. Fang \emph{et al.} \shortcite{fang2022robust} combined FNLL with the heterogeneous model problem and proposed a noise-tolerant loss function and a client confidence re-weighting scheme. It treated all clients equally without explicitly filtering out noisy clients. In contrast, FedCorr \cite{xu2022fedcorr} and FedNoiL \cite{wang2022fednoil} are selection-based methods, which first select noisy clients through designed indicators and then update the federated model by appropriate training strategies. Unfortunately, assuming class-balanced global data makes them less effective when dealing with realistic FL scenarios with class imbalance.
	
	\section{Problem Definition}
	
	\subsection{Preliminaries}
	
	Given a multi-class image classification task where $x_i \in \mathcal{X}$ is the input image and $y_i \in \mathcal{Y} = [C]$ is the corresponding ground-true label. The goal is to use a training dataset $D_{tr} = \{(x_i,y_i)\}_{i=1}^{N}$ of size $N$ which is assumed to be class-imbalanced (\textit{i.e.} $p_{tr}(y) \neq \frac{1}{C}$) to train a model $f(\cdot)$ for classification. In FL, $D_{tr}$ can not be built directly due to privacy concerns, and it is divided into $K$ non-overlapped subsets distributed over $K$ clients. For any private dataset $D_k=\{(x_i,y_i)\}_{i=1}^{N_k}$ of size $N_k$ in client $k \in [K]$, each image-label pair $(x_i, y_i)$ obeys the private joint probability distribution $p_k(x,y)=p_k(y)p_k(x \mid y)$. Following \cite{shen2022agnostic,xu2022fedcorr}, we assume the conditional distribution $p_k(x \mid y)$ is identical, while the local class prior $p_k(y)$ varies across clients (\textit{i.e.} $p_{k_1}(y) \neq p_{k_2}(y)$ for $k_1 \neq k_2$) due to data heterogeneity. To simulate noisy labels, the ground-true label $y$ in some clients is replaced by a noisy label $\overline{y}$.

	\subsection{Noise Model} \label{noise}
	
	\begin{figure}[!t]
		\centering
		\includegraphics[width=0.85\columnwidth]{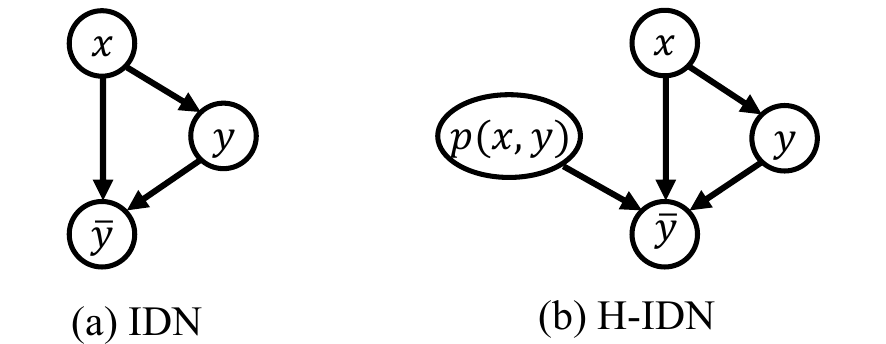}
		\caption{
			Graphical models of standard IDN and the proposed H-IDN.
		}
		\label{fig:noise}
	\end{figure}
	To realistically model label noise in multi-source data, we construct a heterogeneous label noise model. In terms of the strength of label noise, we first define the global noise rate $\rho$ as the proportion of noisy clients. Then, we assume local noise rate $\eta_i$ (\textit{i.e.}, the proportion of noisy samples) follows the uniform distribution $U(\eta^{l}, \eta^{u})$ for any noisy client $i$ as \cite{xu2022fedcorr}. For the pattern of label noise, we propose heterogeneous instance-dependent noise (H-IDN) for better modeling. As shown in Fig. \ref{fig:noise} and Definition \ref{definition1}, local data distribution is an additional factor to manipulate label noise compared to standard instance-dependent noise (IDN) \cite{chen2021beyond}.
	\begin{definition}[H-IDN] \label{definition1}
		IDN is heterogeneous if noise transition probability is a function of local data distribution, \textit{i.e.}, ${\rm Pr}( \overline{Y}=\overline{y} \mid Y=y, X=x)=M_{\overline{y},y,x}(p_i(x,y))$, where $M_{\overline{y},y,x}$ denotes the noise transition matrix of instance $x$.
	\end{definition}
	In this way, the proposed H-IDN is capable of modeling more complicated and realistic label noise.
	
	\begin{algorithm}[tb]
		\caption{Noise Generation.}
		\label{alg:NG}
		\textbf{Input}: Number of clients $K$; clean local datasets $\{D_k\}_{k=1}^K$; global noise rate $\rho$; local noise rate distribution parameters $\eta^l, \eta^u$.
		\begin{algorithmic}[1] 
			\STATE $\mathcal{I} = $ Randomly select $\rho K$ elements from $[K]$.
			\FOR{$i$ in $\mathcal{I}$}
			\STATE Initialize a network $f_i$.
			\STATE Train $f_i$ on the local dataset $D_i = \{(x_j,y_j)\}_{j=1}^{N_i}$.
			\STATE Compute classification probabilities $p(Y \mid x) \in [0, 1]^{N_i \times C}$ for all samples in $D_i$.
			\STATE Compute the misclassification probability $\widetilde{p}(x) \in [0, 1]^{N_i}$ for each sample in $D_i$ (Eq. \ref{mis}).
			\STATE $\eta_i \sim U(\eta^l,\eta^u)$. 
			\STATE $\mathcal{N} = $ Randomly select $\eta_i N_i$ elements from $[N_i]$ with the probability $\widetilde{p}(x) / \sum \widetilde{p}(x)$. \hfill \textcolor{blue}{$\triangleright$ \textit{Normalization}}
			\FOR{$t$ in $\mathcal{N}$}
			\STATE $\overline{y}_t = $ Randomly select a different label from $\mathcal{Y}$ with the probability $p(Y \mid x_t)$.
			\STATE Flip $y_t$ to $\overline{y}_t$. \hfill \textcolor{blue}{$\triangleright$ \textit{Add label noise}}
			\ENDFOR
			\ENDFOR
		\end{algorithmic}
		\textbf{Output}: Local datasets after adding label noise $\{D_k\}_{k=1}^K$. 
	\end{algorithm}
	
	\begin{figure*}[!t]
		\centering
		\includegraphics[width=1\textwidth]{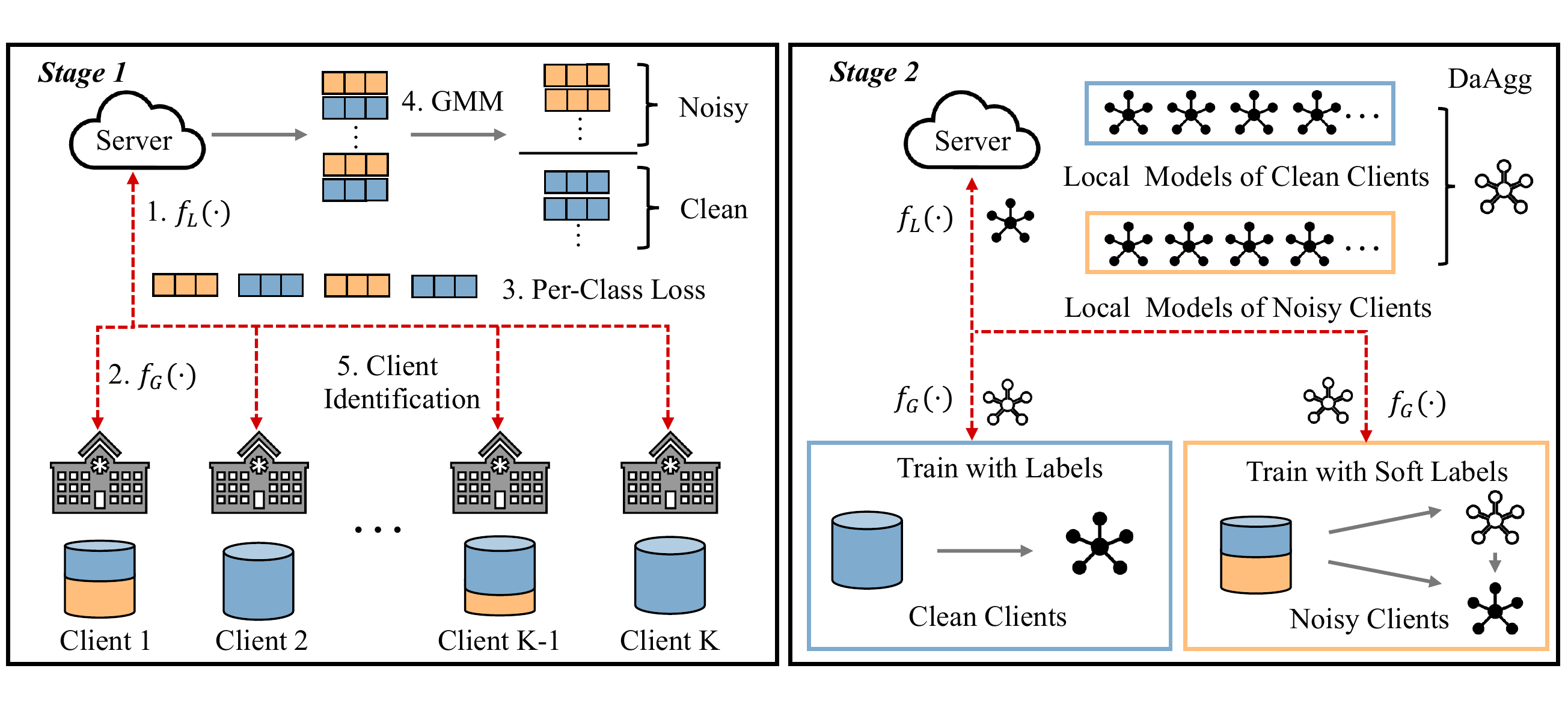}
		\caption{Overview of the proposed two-stage framework FedNoRo.}
		\label{framework}
	\end{figure*}
	
	\subsection{Noise Generation}
	
	To simulate the noise model in Sec. \ref{noise}, we present a noise generator to produce heterogeneous label noise from multi-source clean data as described in Algorithm \ref{alg:NG}. 
	Following \cite{chen2021beyond}, neural networks are used as annotators to simulate the noisy labeling process in noisy clients. Specifically, we first randomly select $\rho K$ out of $K$ clients as noisy clients and independently train a neural network based on each selected client's original data (\textit{i.e.}, clean data). Then, the trained neural network is used to produce the classification probabilities of all samples belonging to the client. Given each instance $x_j$ in the $i$-th noisy client and the corresponding classification probability $p(Y \mid x_j) \in [0, 1]^C$, its misclassification probability is determined by 
	\begin{equation} \label{mis}
		\widetilde{p}(x_j) = 1 - p(Y=y_j \mid x_j),
	\end{equation}
	and totally $\eta_i N_i$ samples would be chosen as noisy samples based on the normalized misclassification probability $\widetilde{p}(x) \in [0, 1]^{N_i}$. According to Eq. \ref{mis}, hard samples are more likely to be selected as noisy samples. 
	For each selected noisy sample $x_j$, it is assigned a noisy label $\overline{y}$ different from the original clean label $y$ from the probability distribution $p(Y \mid x_j)$. In this way, noisy samples are more likely to be assigned with wrong labels of higher prediction confidence. 
	As the noise of different clients is generated by different neural networks, it fits the definition of a heterogeneous noise pattern. Following the above generation process, generated noisy labels are more reasonable and challenging.
	
	\section{Methodology}
	To address class imbalance and label noise heterogeneity, we propose a two-stage framework named FedNoRo as illustrated in Fig. \ref{framework}. In the first stage, a warm-up model is firstly trained based on FedAvg (\textit{i.e.} steps 1 and 2), which is used to calculate the per-class loss values of each client (\textit{i.e.} step 3). Then, client-wise per-class loss values are collected and processed in the server and used to identify noisy clients by Gaussian Mixture Model (GMM) (\textit{i.e.} steps 4 and 5). According to the client identification results of the first stage, different training strategies are employed for clean and noisy clients respectively in the second stage. Specifically, the vanilla cross-entropy loss is adopted for clean clients while knowledge distillation (KD) is employed for better noisy label learning for noisy clients. Moreover, a distance-aware model aggregation function is defined to reduce the negative impact of noisy clients on global model updating. It should be noted that, in both stages, logit adjustment (LA) \cite{menon2020long} is introduced to local training to combat data heterogeneity and class imbalance.
	
	\subsection{Stage 1: Noisy Client Detection}
	As discussed in Section \ref{section1}, class imbalance destroys the law that the client-wise average indicator (\textit{e.g.} loss value) of clean clients is independent and identically distributed (IID), causing failure to identify noisy clients with client-wise indicators. To address this, the key is to re-find an IID indicator for clean clients that can detect noisy clients as outliers.
	
	For the identification of noisy clients, we first train a warm-up model for $T_1$ (will be discussed in \ref{t1}) rounds by FedAvg 
	\begin{equation} \label{fedavg}
		w_g = \sum_{i=1}^K \frac{N_i}{\sum_{j=1}^K N_j } w_i,
	\end{equation}
	where $w_g$ and $w_i$ denote the weights of the global model and the $i$-th local model respectively and $K$ is the number of clients for aggregation. Then, in the local training phase, \textcolor{black}{vanilla cross-entropy loss is adopted to train the local model combined with LA as $f(\cdot)+\log \pi$}, where $\pi$ represents the local class prior (\textit{i.e.} label distribution). LA is to make the local model treat each class equally rather than being biased, regardless of the biased class prior in local data (\textit{i.e.} local class imbalance). In this way, different local models with data heterogeneity are more homogeneous by pursuing a unified learning goal regardless of their biased local class priors, which in turn helps model aggregation. Furthermore, as all local models participating in the aggregation are unbiased, the global model is more likely to be unbiased. However, such an unbiased global model is proved to be unsuitable for noisy client identification through client-wise average indicators \cite{xu2022fedcorr,wang2022fednoil}. It is because, though LA is helpful to alleviate the bias in class priors, the bias in learning difficulty of class-specific features is unsolved \cite{yi2022identifying}.
	
	To complement this, per-class indicators are calculated, instead of using an averaged indicator for samples of all classes. It is based on our observation in clinical practice that, though inter-class samples are non-IID under class imbalance, intra-class samples are more IID. That's because medical imaging data is captured strictly following standard protocols. Given certain diseases, patients exhibit similar symptoms, and the imaging equipment required for diagnosis is strictly fixed, leading to images with the same or similar styles even at different hospitals. In other words, the IID assumption is satisfied by separately focusing on each specific class $c$ on each client's private data. Therefore, given any class $c$, clients containing noisy samples in class $c$ can be effectively identified as outliers (\textit{e.g.}, larger loss values). In addition, under label noise heterogeneity, noisy samples across clients may be distributed to different noisy labels, and relying on any single class indicator is insufficient to identify noisy clients. Thus, the average loss values of all classes on each client $i$ denoted as $l_i=(l_i^1,l_i^2,...,l_i^C)^T \in \mathbb{R}^C$ are used for identification. 
	
	Given $K$ per-class average loss values $L=[l_1,...,l_K]$, noisy client detection is implemented in an unsupervised learning way. Considering the class-missing problem in heterogeneous data, a specific class $c$ may not exist in client $i$, leading to a missing average loss value $l_i^c$ in $l_i$. Such missing values in $L$ cannot be used for unsupervised learning and must be filled. Actually, if class $c$ is missing in client $i$, it will surely have no noisy samples in $c$. According to the small-loss trick \cite{han2018co}, the cleaner a sample is, the easier it is to learn and the smaller its loss is. Therefore, $l_i^c$ is simply replaced by the minimum value of class $c$ across all clients (\textit{i.e.}, $\min_{i=1}^K l_i^c$). 
	Given the bias of learning difficulty of different classes, loss value ranges across classes can vary dramatically, and the average loss values of all clients for any class should be uniformly normalized to $[0,1]$ by  
	\begin{equation} \label{minmax}
		l_i^c = \frac{l_i^c-\min_i l_i^c}{\max_i l_i^c-\min_i l_i^c}.
	\end{equation}
	To this end, all classes contribute equally to identifying noisy clients instead of being dominated by certain classes. Then, a two-component Gaussian Mixture Model (GMM) is deployed onto the processed client-wise per-class loss vectors, and $K$ clients are partitioned into two subsets: $S_c$ (\textit{i.e.}, clean clients corresponding to the Gaussian distribution with a small norm mean vector) and $S_n$ (\textit{i.e.}, noisy clients corresponding to the Gaussian distribution with a large norm mean vector).
	It should be noted that this noisy client detection process is privacy-preserving as only average loss values are uploaded to the server.
	
	\subsection{Stage 2: Noise-Robust Training}
	
	After noisy client detection, customized training and aggregation strategies are proposed for clean and noisy clients respectively for noise-robust federated learning.
	Similar to the first stage, in the local training phase, LA is introduced to both clean and noisy clients to address data heterogeneity and global class imbalance. For clean clients, the vanilla cross-entropy loss is adopted to train each local model based on clean labels. For noisy clients, inspired by the label-soften-based method which has shown great potential in noisy label learning \cite{wei2021smooth,lukasik2020does,lukov2022teaching}, a knowledge distillation (KD) \cite{hinton2015distilling} based training method is applied. More specifically, given any $x$ and its output logit from the global model $f_G(x)$, a targeted probability distribution is calculated as
	\begin{equation}
		y_G = {\rm softmax}(\frac{f_G(x)}{T}),
	\end{equation}
	where $T$ is the temperature to control the uniformness of $y_G$ and is set as 0.8 by default in this paper. Then, the total training loss for each local model in noisy clients is written as
	\begin{equation}
		\mathcal{L} = \lambda \mathcal{L}_{KL}(y_p, y_G) + (1-\lambda) \mathcal{L}_{CE}(y_p, \overline{y}),
	\end{equation}
	where $y_p$ represents the prediction results of the local model, $\mathcal{L}_{KL}$ is the Kullback-Leibler divergence, and $\lambda$ is a trade-off coefficient. Inspired by semi-supervised learning \cite{laine2016temporal}, $\lambda$ grows from 0 to $\lambda_{max}$ (set as 0.8 by default) using a Gaussian ramp-up curve. Considering the memorization effect of neural networks on noisy datasets \cite{han2018co,jiang2022towards}, this strategy would be robust for noisy clients. In the early training stage, each local model tends to learn simple patterns (more likely to be clean samples), and the training process is dominated by the hard labels $\overline{y}$ and the cross entropy loss. As the neural network begins to fit complex patterns (more likely to be noisy samples), soft labels $y_G$ gradually dominate the training process to avoid the model overfitting to noisy labels.
	
	In the model aggregation phase, models from clean and noisy clients are of different importance. Hence, inspired by \cite{liang2022rscfed}, distance-aware model aggregation DaAgg is proposed where a client-wise distance metric is defined as
	\begin{equation}
		d(i) = \min_{j \in S_c} \lVert w_i - w_j \rVert_2,
	\end{equation}
	$w_i$ denotes the weights of the $i$-th local model. This metric measures the distance between a model $w_i$ and the nearest model of clean clients. Note that $d(i)$ is equal to 0 if client $i$ is clean. To ensure the boundness of $d(i)$, it is further normalized to $[0,1]$ as $D(i) = \frac{d(i)}{\max_j d(j)}$. Then, local models are aggregated to update the global model by
	\begin{equation} \label{agg}
		w_g = \sum_{i=1}^K \frac{N_i e^{-D(i)}}{\sum_{j=1}^K N_j e^{-D(j)}} w_i.
	\end{equation}
	In this way, the aggregation weight of any clean client $i$ is constant due to $D(i)=0$, while the weights of noisy clients are multiplied by a scaling factor in $[0,1]$. Moreover, the weights would decrease as the model distance increases, adaptively adjusting the contributions of noisy clients in global model updating for noise-robust federated learning.
	
	\section{Experimental Results}
	\subsection{Dataset and Evaluation Metric}
	Two widely-used datasets are included for evaluation:
	\begin{enumerate}
		\item ICH dataset \cite{flanders2020construction} consisting of 67969 brain CT slices for the diagnosis of five intracranial hemorrhage (ICH) subtypes. The dataset is pre-processed following \cite{liu2021federated,jiang2022dynamic}.
		
		\item ISIC 2019 dataset \cite{tschandl2018ham10000,codella2018skin,combalia2019bcn20000} containing 25331 images for the classification of eight skin diseases.
	\end{enumerate}
	Both datasets are randomly divided into the training and test sets following a 7:3 split. As both datasets are highly class-imbalanced, the divided training and test sets are also class-imbalanced, satisfying the proposed setting. Considering the imbalanced test sets, the official ISIC Challenge 2019 metric, balanced accuracy (BACC), is adopted for evaluation.
	
	\subsection{Experimental Setup} \label{es}
	
	\subsubsection{Data Partition and Noise Generation}
	
\begin{figure}[!t] 
		\centering
		\subfigure[ICH]{\includegraphics[width=0.48\columnwidth]{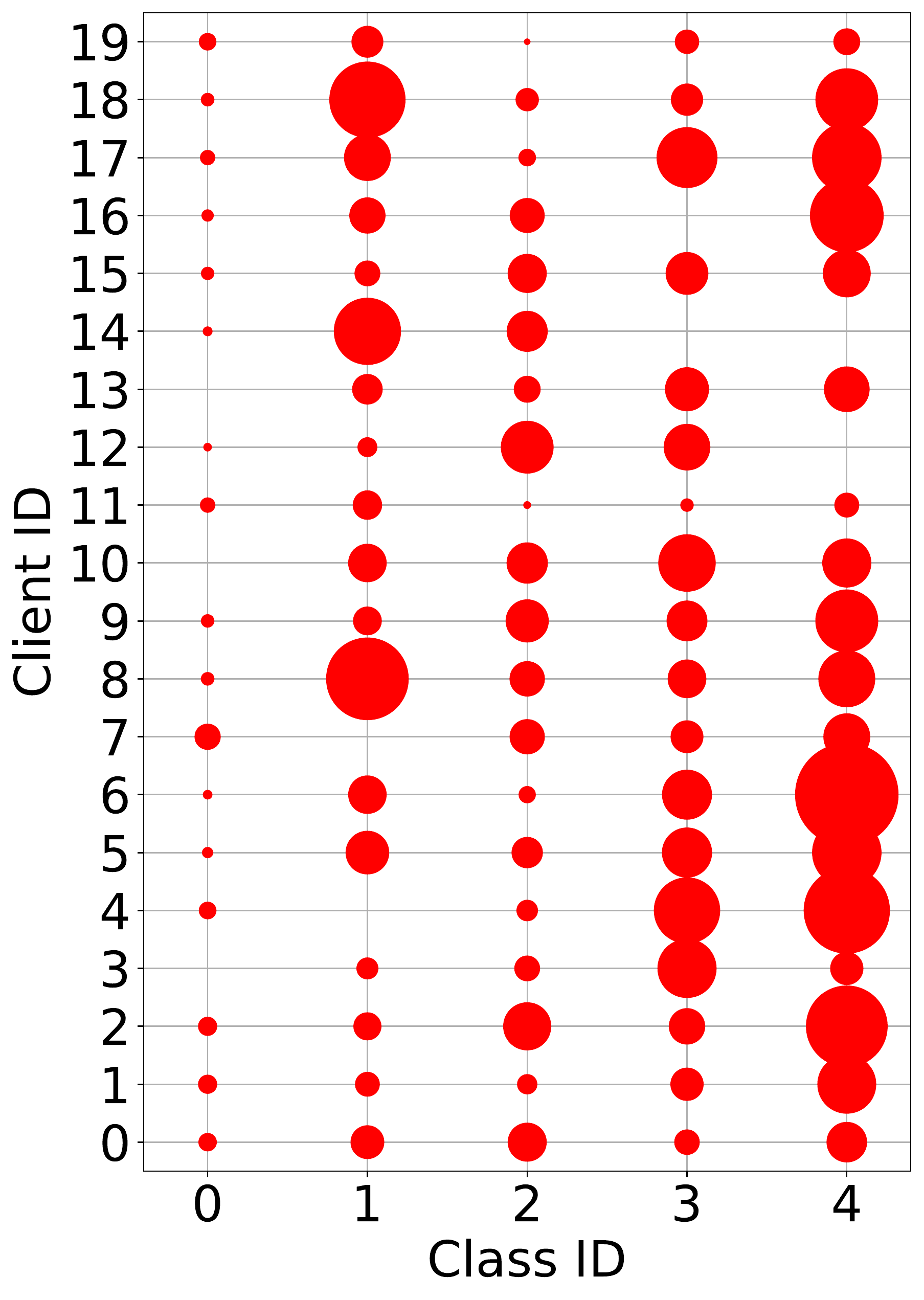}}
		\subfigure[ISIC 2019]{\includegraphics[width=0.48\columnwidth]{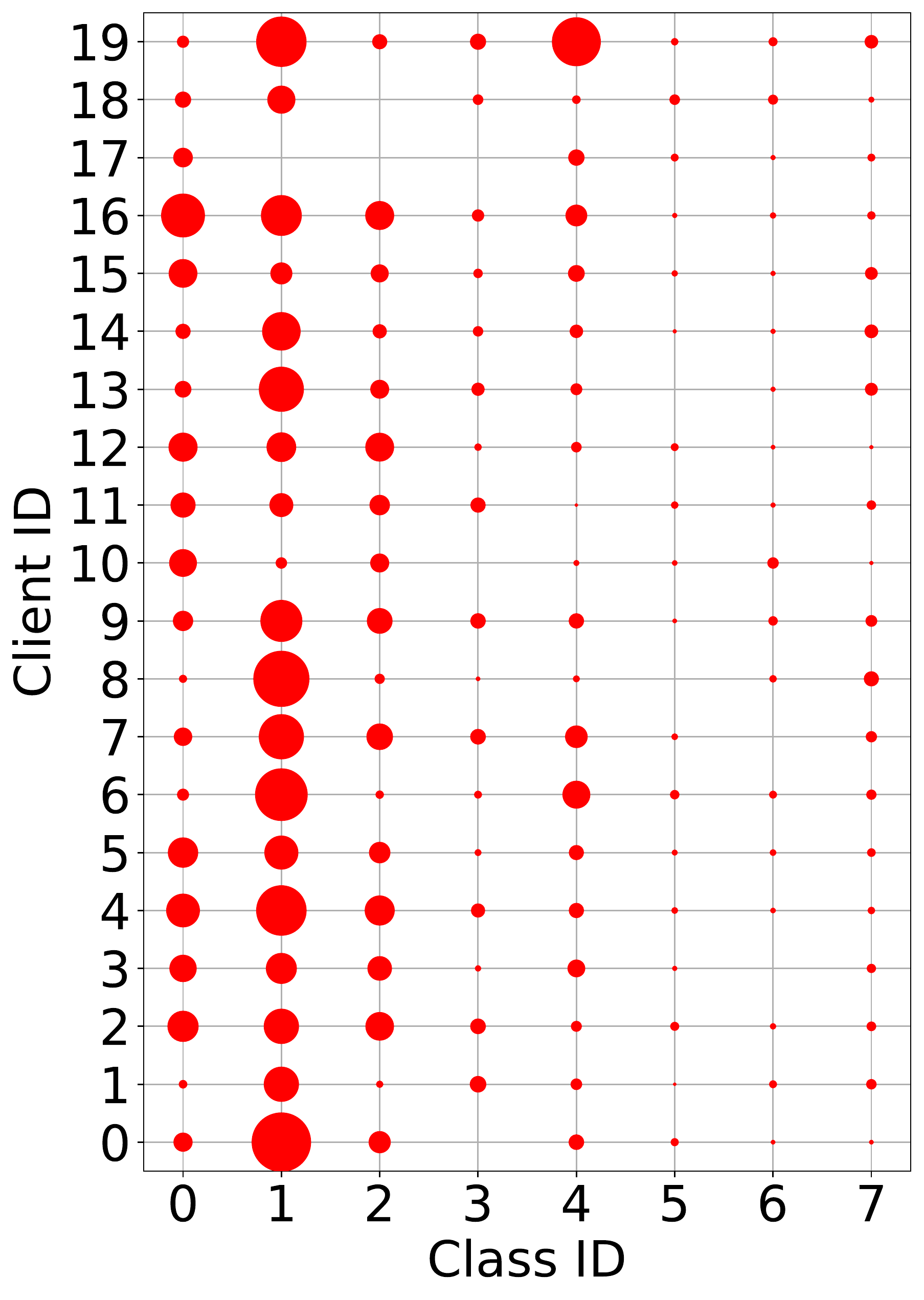}}
		\caption{Visualization of data distributions of the two datasets.}
		\label{datadistribution}
	\end{figure}

	Following the previous work on FNLL \cite{xu2022fedcorr}, we use Bernoulli distribution and Dirichlet distribution to generate heterogeneous data partitions. Specifically, we set $\alpha_{Dir}=2.0, \Phi_{ij} \sim {\rm Bernoulli}(0.9), K=20$ and $\alpha_{Dir}=1.5, \Phi_{ij} \sim {\rm Bernoulli}(0.99), K=20$ for the ICH and ISIC 2019 datasets respectively. For better visualization, we use dot size to represent the amount of data and plot the data distributions as shown in Fig. \ref{datadistribution}.
	For evaluation under various settings of label noise, $\rho$ is set as 0.2, 0.3, 0.4, and 0.6, and ($\eta^l,\eta^u$) is set as (0.3,0.5) and (0.5,0.7) for the ICH dataset, while $\rho$ is set as 0.4 and 0.6 and ($\eta^l,\eta^u$) is set as (0.5,0.7) for the ISIC 2019 dataset.
	\begin{table*}[t]
		\centering
		\renewcommand{\arraystretch}{0.8}
		\resizebox{\textwidth}{!}{
			\begin{tabular}{llcccccccccc}
				\toprule
				\hline
				\multicolumn{1}{l|}{\multirow{2}{*}{Category}}   & \multicolumn{1}{l|}{\multirow{2}{*}{Methods}}    & \multicolumn{1}{c|}{$\rho$}             & \multicolumn{1}{c|}{0.0}            & \multicolumn{2}{c|}{0.2}                                                  & \multicolumn{2}{c|}{0.3}                                                  & \multicolumn{2}{c|}{0.4}                                                  & \multicolumn{2}{c}{0.6}                              \\ \cline{3-12} 
				\multicolumn{1}{l|}{}                            & \multicolumn{1}{l|}{}                            & \multicolumn{1}{c|}{($\eta^l,\eta^u$)} & \multicolumn{1}{c|}{$(0.0,0.0)$}      & \multicolumn{1}{c|}{$(0.3,0.5)$}      & \multicolumn{1}{c|}{($0.5,0.7)$}      & \multicolumn{1}{c|}{$(0.3,0.5)$}      & \multicolumn{1}{c|}{$(0.5,0.7)$}      & \multicolumn{1}{c|}{$(0.3,0.5)$}      & \multicolumn{1}{c|}{$(0.5,0.7)$}      & \multicolumn{1}{c|}{$(0.3,0.5)$}      & $(0.5,0.7)$      \\ \hline
				\multicolumn{1}{l|}{\multirow{6}{*}{FL}}         & \multicolumn{1}{l|}{\multirow{2}{*}{FedAvg}}     & \multicolumn{1}{c|}{Best}              & \multicolumn{1}{c|}{69.34}          & \multicolumn{1}{c|}{65.81}          & \multicolumn{1}{c|}{64.53}          & \multicolumn{1}{c|}{62.49}          & \multicolumn{1}{c|}{60.82}          & \multicolumn{1}{c|}{60.52}          & \multicolumn{1}{c|}{58.38}          & \multicolumn{1}{c|}{58.46}          & 54.77          \\
				\multicolumn{1}{l|}{}                            & \multicolumn{1}{l|}{}                            & \multicolumn{1}{c|}{Last}              & \multicolumn{1}{c|}{68.92}          & \multicolumn{1}{c|}{65.33}          & \multicolumn{1}{c|}{63.97}          & \multicolumn{1}{c|}{62.10}          & \multicolumn{1}{c|}{60.23}          & \multicolumn{1}{c|}{60.05}          & \multicolumn{1}{c|}{56.79}          & \multicolumn{1}{c|}{56.88}          & 50.35          \\ \cline{2-12} 
				\multicolumn{1}{l|}{}                            & \multicolumn{1}{l|}{\multirow{2}{*}{FedProx}}    & \multicolumn{1}{c|}{Best}              & \multicolumn{1}{c|}{68.16}          & \multicolumn{1}{c|}{64.58}          & \multicolumn{1}{c|}{63.64}          & \multicolumn{1}{c|}{61.81}          & \multicolumn{1}{c|}{61.91}          & \multicolumn{1}{c|}{60.85}          & \multicolumn{1}{c|}{58.50}          & \multicolumn{1}{c|}{60.21}          & 57.57          \\
				\multicolumn{1}{l|}{}                            & \multicolumn{1}{l|}{}                            & \multicolumn{1}{c|}{Last}              & \multicolumn{1}{c|}{67.29}          & \multicolumn{1}{c|}{63.80}          & \multicolumn{1}{c|}{62.65}          & \multicolumn{1}{c|}{61.46}          & \multicolumn{1}{c|}{61.00}          & \multicolumn{1}{c|}{59.99}          & \multicolumn{1}{c|}{57.85}          & \multicolumn{1}{c|}{59.35}          & 56.63          \\ \cline{2-12} 
				\multicolumn{1}{l|}{}                            & \multicolumn{1}{l|}{\multirow{2}{*}{FedLA}}      & \multicolumn{1}{c|}{Best}              & \multicolumn{1}{c|}{\textbf{73.56}} & \multicolumn{1}{c|}{69.45}          & \multicolumn{1}{c|}{69.28}          & \multicolumn{1}{c|}{66.84}          & \multicolumn{1}{c|}{64.84}          & \multicolumn{1}{c|}{66.60}          & \multicolumn{1}{c|}{62.39}          & \multicolumn{1}{c|}{63.90}          & 58.78          \\
				\multicolumn{1}{l|}{}                            & \multicolumn{1}{l|}{}                            & \multicolumn{1}{c|}{Last}              & \multicolumn{1}{c|}{\textbf{73.07}} & \multicolumn{1}{c|}{68.82}          & \multicolumn{1}{c|}{68.45}          & \multicolumn{1}{c|}{66.20}          & \multicolumn{1}{c|}{64.03}          & \multicolumn{1}{c|}{63.90}          & \multicolumn{1}{c|}{60.51}          & \multicolumn{1}{c|}{61.86}          & 54.99          \\ \hline
				\multicolumn{1}{l|}{\multirow{8}{*}{Denoise FL}} & \multicolumn{1}{l|}{\multirow{2}{*}{RoFL}}       & \multicolumn{1}{c|}{Best}              & \multicolumn{1}{c|}{-}              & \multicolumn{1}{c|}{42.57}          & \multicolumn{1}{c|}{40.42}          & \multicolumn{1}{c|}{40.95}          & \multicolumn{1}{c|}{39.87}          & \multicolumn{1}{c|}{40.35}          & \multicolumn{1}{c|}{35.60}          & \multicolumn{1}{c|}{40.18}          & 35.47          \\
				\multicolumn{1}{l|}{}                            & \multicolumn{1}{l|}{}                            & \multicolumn{1}{c|}{Last}              & \multicolumn{1}{c|}{-}              & \multicolumn{1}{c|}{42.19}          & \multicolumn{1}{c|}{40.30}          & \multicolumn{1}{c|}{40.64}          & \multicolumn{1}{c|}{39.27}          & \multicolumn{1}{c|}{40.25}          & \multicolumn{1}{c|}{35.39}          & \multicolumn{1}{c|}{39.68}          & 35.35          \\ \cline{2-12} 
				\multicolumn{1}{l|}{}                            & \multicolumn{1}{l|}{\multirow{2}{*}{RHFL}}       & \multicolumn{1}{c|}{Best}              & \multicolumn{1}{c|}{-}              & \multicolumn{1}{c|}{57.48}          & \multicolumn{1}{c|}{56.91}          & \multicolumn{1}{c|}{56.72}          & \multicolumn{1}{c|}{55.74}          & \multicolumn{1}{c|}{55.26}          & \multicolumn{1}{c|}{54.30}          & \multicolumn{1}{c|}{54.63}          & 51.00          \\
				\multicolumn{1}{l|}{}                            & \multicolumn{1}{l|}{}                            & \multicolumn{1}{c|}{Last}              & \multicolumn{1}{c|}{-}              & \multicolumn{1}{c|}{57.05}          & \multicolumn{1}{c|}{56.46}          & \multicolumn{1}{c|}{55.75}          & \multicolumn{1}{c|}{52.19}          & \multicolumn{1}{c|}{54.71}          & \multicolumn{1}{c|}{52.08}          & \multicolumn{1}{c|}{52.53}          & 49.64          \\ \cline{2-12} 
				\multicolumn{1}{l|}{}                            & \multicolumn{1}{l|}{\multirow{2}{*}{FedLSR}}     & \multicolumn{1}{c|}{Best}              & \multicolumn{1}{c|}{-}              & \multicolumn{1}{c|}{55.99}          & \multicolumn{1}{c|}{55.28}          & \multicolumn{1}{c|}{54.44}          & \multicolumn{1}{c|}{52.27}          & \multicolumn{1}{c|}{52.48}          & \multicolumn{1}{c|}{48.20}          & \multicolumn{1}{c|}{50.89}          & 43.30          \\
				\multicolumn{1}{l|}{}                            & \multicolumn{1}{l|}{}                            & \multicolumn{1}{c|}{Last}              & \multicolumn{1}{c|}{-}              & \multicolumn{1}{c|}{55.69}          & \multicolumn{1}{c|}{54.77}          & \multicolumn{1}{c|}{53.95}          & \multicolumn{1}{c|}{51.70}          & \multicolumn{1}{c|}{51.92}          & \multicolumn{1}{c|}{46.77}          & \multicolumn{1}{c|}{49.96}          & 39.81          \\ \cline{2-12} 
				\multicolumn{1}{l|}{}                            & \multicolumn{1}{l|}{\multirow{2}{*}{FedCorr}}    & \multicolumn{1}{c|}{Best}              & \multicolumn{1}{c|}{-}              & \multicolumn{1}{c|}{57.90}          & \multicolumn{1}{c|}{56.68}          & \multicolumn{1}{c|}{55.02}          & \multicolumn{1}{c|}{54.61}          & \multicolumn{1}{c|}{53.62}          & \multicolumn{1}{c|}{50.82}          & \multicolumn{1}{c|}{50.89}          & 47.13          \\
				\multicolumn{1}{l|}{}                            & \multicolumn{1}{l|}{}                            & \multicolumn{1}{c|}{Last}              & \multicolumn{1}{c|}{-}              & \multicolumn{1}{c|}{57.68}          & \multicolumn{1}{c|}{55.86}          & \multicolumn{1}{c|}{54.52}          & \multicolumn{1}{c|}{53.62}          & \multicolumn{1}{c|}{52.60}          & \multicolumn{1}{c|}{49.59}          & \multicolumn{1}{c|}{50.24}          & 46.13          \\ \hline
				\multicolumn{1}{l|}{\multirow{2}{*}{Joint}}      & \multicolumn{1}{l|}{\multirow{2}{*}{FedCorr+LA}} & \multicolumn{1}{c|}{Best}              & \multicolumn{1}{c|}{-}              & \multicolumn{1}{c|}{64.78}          & \multicolumn{1}{c|}{64.29}          & \multicolumn{1}{c|}{62.58}          & \multicolumn{1}{c|}{62.78}          & \multicolumn{1}{c|}{62.19}          & \multicolumn{1}{c|}{60.06}          & \multicolumn{1}{c|}{57.88}          & 54.53          \\
				\multicolumn{1}{l|}{}                            & \multicolumn{1}{l|}{}                            & \multicolumn{1}{c|}{Last}              & \multicolumn{1}{c|}{-}              & \multicolumn{1}{c|}{63.99}          & \multicolumn{1}{c|}{63.55}          & \multicolumn{1}{c|}{61.90}          & \multicolumn{1}{c|}{61.87}          & \multicolumn{1}{c|}{61.30}          & \multicolumn{1}{c|}{59.58}          & \multicolumn{1}{c|}{57.17}          & 53.92          \\ \hline
				\multicolumn{1}{l|}{\multirow{2}{*}{Ours}}       & \multicolumn{1}{l|}{\multirow{2}{*}{FedNoRo}}        & \multicolumn{1}{c|}{Best}              & \multicolumn{1}{c|}{-}              & \multicolumn{1}{c|}{\textbf{70.59}} & \multicolumn{1}{c|}{\textbf{70.64}} & \multicolumn{1}{c|}{\textbf{70.14}} & \multicolumn{1}{c|}{\textbf{69.35}} & \multicolumn{1}{c|}{\textbf{70.69}} & \multicolumn{1}{c|}{\textbf{69.30}} & \multicolumn{1}{c|}{\textbf{67.55}} & \textbf{63.83} \\
				\multicolumn{1}{l|}{}                            & \multicolumn{1}{l|}{}                            & \multicolumn{1}{c|}{Last}              & \multicolumn{1}{c|}{-}              & \multicolumn{1}{c|}{\textbf{70.18}} & \multicolumn{1}{c|}{\textbf{69.81}} & \multicolumn{1}{c|}{\textbf{69.29}} & \multicolumn{1}{c|}{\textbf{68.47}} & \multicolumn{1}{c|}{\textbf{70.14}} & \multicolumn{1}{c|}{\textbf{68.87}} & \multicolumn{1}{c|}{\textbf{67.10}} & \textbf{63.29} \\ \hline
				\bottomrule
			\end{tabular}
		}
		\caption{Quantitative BACC (\%) comparison results on the ICH dataset under different noise rates. The best results are marked in bold.}
		\label{tab:ich}
	\end{table*}
	
	\subsubsection{Implementation Details}
	
	\begin{table}[t]
		\centering
		\renewcommand{\arraystretch}{0.8}
		\resizebox{\columnwidth}{!}{
			\begin{tabular}{l|c|c|c|c}
				\toprule
				\hline
				\multirow{2}{*}{Methods}    & $\rho$           & 0.0            & 0.4            & 0.6            \\ \cline{2-5} 
				& ($\eta^l,\eta^u$) & $(0.0,0.0)$      & $(0.5,0.7)$      & $(0.5,0.7)$      \\ \hline
				\multirow{2}{*}{FedAvg}     & Best            & 69.09          & 61.83          & 54.35          \\
				& Last            & 65.44          & 58.23          & 49.88          \\ \hline
				\multirow{2}{*}{FedProx}    & Best            & 72.20          & 64.46          & 58.55          \\
				& Last            & \textbf{69.60} & 60.13          & 49.87          \\ \hline
				\multirow{2}{*}{FedLA}      & Best            & \textbf{72.55} & 66.34          & 61.20          \\
				& Last            & 68.72          & 61.18          & 56.11          \\ \hline
				\multirow{2}{*}{RoFL}       & Best            & -              & 28.45              & 28.86          \\
				& Last            & -              & 27.79              & 28.29          \\ \hline
				\multirow{2}{*}{RHFL}       & Best            & -              & 46.06          & 46.67          \\
				& Last            & -              & 44.04          & 45.09          \\ \hline
				\multirow{2}{*}{FedLSR}     & Best            & -              & 30.15          & 27.24          \\
				& Last            & -              & 29.11          & 26.08          \\ \hline
				\multirow{2}{*}{FedCorr}    & Best            & -              & 42.54          & 38.40          \\
				& Last            & -              & 41.12          & 37.17          \\ \hline
				\multirow{2}{*}{FedCorr+LA} & Best            & -              & 60.38          & 55.40          \\
				& Last            & -              & 59.16          & 54.27          \\ \hline
				\multirow{2}{*}{FedNoRo (ours)}        & Best            & -              & \textbf{68.59} & \textbf{66.00} \\
				& Last            & -              & \textbf{64.67} & \textbf{60.65} \\ \hline
				\bottomrule
			\end{tabular}
		}
		\caption{Quantitative BACC (\%) results on the ISIC 2019 dataset under different noise rates. The best results are marked in bold.}
		\label{tab:isic}
	\end{table}
	ResNet-18 \cite{he2016deep} is used as the backbone model for the ICH and ISIC 2019 dataset (pre-trained by ImageNet). In terms of federated settings, the maximal communication round is set as 100 and the local epoch is set as 5 for the ICH dataset while set as 300 and 1 respectively for the ISIC 2019 dataset. $T_1$ for warm-up training is set to 10 and 20 on the two datasets respectively. Other settings are the same for both datasets, including an Adam optimizer with momentum terms of (0.9, 0.999), a weight decay of 5e-4, a constant learning rate of 3e-4, and a batch size of 16.
	
	\subsection{Comparison with SOTA Methods}
	
	A series of state-of-the-art methods are selected for comparison, including conventional FL methods (\textit{i.e.}, FedAvg \cite{mcmahan2017communication}, FedProx \cite{FedProx}, and FedLA \cite{menon2020long}) and de-noise FL methods (\textit{i.e.}, RoFL \cite{yang2022robust}, RHFL \cite{fang2022robust}, FedLSR \cite{jiang2022towards}, and FedCorr \cite{xu2022fedcorr}). We also combine LA with FedCorr for comparison.
	
	Quantitative results, including both the best BACC across all rounds (denoted as Best) and the averaged BACC over the last 10 rounds (denoted as Last), on the two datasets are summarized in Tabs. \ref{tab:ich} and \ref{tab:isic}. FedLA generally achieves the best performance among conventional FL methods, benefiting from alleviating class imbalance. Meanwhile, all conventional FL methods would suffer from performance degradation with the increase in noise rate. Unfortunately, though designed for noisy label learning, the four de-noise FL methods fail to achieve performance improvement. Without a specific solution to class imbalance, label noise learning will be seriously affected by class imbalance, resulting in even worse performance compared to the baseline. It explains why introducing LA to FedCorr is beneficial. But as analyzed before, class imbalance cannot be completely eliminated by LA, simply combing LA with FedCorr is sub-optimal. Comparatively, FedNoRo consistently outperforms other methods by a large margin. Furthermore, the performance of FedNoRo is more stable with the increase in noise rate, demonstrating the noise-robust nature of FedNoRo.
	
	\subsection{Ablation Study}
	
	\subsubsection{How Each Step in Noisy Client Detection Works}
	
	Additional experiments are conducted on the ICH dataset to analyze how each step in the first stage of FedNoRo affects noisy client detection as summarized in Tab. \ref{tab:ablation1}.
	Here, \textit{Per-Class} is to calculate certain indicators separately for each class, and \textit{Norm.} is to apply min-max normalization to the \textit{Per-Class} indicator according to Eq. \ref{minmax}. 
	For a fair comparison, the number of training rounds in the first stage of all methods is the same as Sec. \ref{es}. Given certain client-wise indicators under a specific setting, we repeatedly adopt GMM (10000 different random seeds) to identify noisy clients. Then, the average recall (\textit{i.e.}, Re), precision (\textit{i.e.}, Pr), and matching ratio (\textit{i.e.}, MR: ratio of successfully-identified noisy clients) of 10,000 times of identifications are used to measure whether the method can effectively detect noisy clients. 
	
	As stated in Tab. \ref{tab:ablation1}, the first two methods from \cite{xu2022fedcorr} and \cite{wang2022fednoil}, together with the third one, completely ignore class imbalance, resulting in poor detection results. Introducing LA to address class imbalance is somewhat helpful for noisy client detection, leading to slight performance improvement. But it fails to improve the MR performance. Comparatively, when each class is considered independently, both noise heterogeneity and class imbalance are further alleviated, leading to better MR performance. When combing LA and \textit{Per-Class} with min-max normalization, namely FedNoRo, most noisy clients are effectively detected, which lays a solid foundation for the second stage.
	\begin{table}[t]
		\centering
		\renewcommand{\arraystretch}{0.6}
		\resizebox{\columnwidth}{!}{
			\begin{tabular}{cccc|ccc}
				\toprule
				\hline
				\multicolumn{1}{c|}{\textit{Indicator}}  & LA           & \textit{Per-Class}    & \textit{Norm.}   & Re (\%)              & Pr (\%)           & MR (\%)               \\ \hline
				\multicolumn{7}{c}{ICH, $\rho=0.3$, $(\eta^l,\eta^u)=(0.3,0.5)$} \\ \hline
				\multicolumn{1}{c|}{LID}        &\XSolidBrush              &\XSolidBrush              &\XSolidBrush              & \textbf{100.00}     & 54.54               & 0.00               \\
				\multicolumn{1}{c|}{Conf.} &\XSolidBrush              &\XSolidBrush              &\XSolidBrush              & 16.66               & 8.33                & 0.00               \\
				\multicolumn{1}{c|}{Loss}       &\XSolidBrush              &\XSolidBrush              &\XSolidBrush              & 94.58               & 57.12               & 0.00               \\
				\multicolumn{1}{c|}{Loss}       & \Checkmark &\XSolidBrush              &\XSolidBrush              & \textbf{100.00}     & 47.14               & 0.00               \\
				\multicolumn{1}{c|}{Loss}       &\XSolidBrush              & \Checkmark &\XSolidBrush              & 97.19               & 90.63               & 85.77              \\
				\multicolumn{1}{c|}{Loss}       & \Checkmark & \Checkmark &\XSolidBrush              & 99.48               & 97.78               & 96.93              \\
				\multicolumn{1}{c|}{Loss}       & \Checkmark & \Checkmark & \Checkmark & 99.70               & \textbf{98.76}      & \textbf{98.28}     \\ \hline
				\multicolumn{7}{c}{ICH, $\rho=0.4$, $(\eta^l,\eta^u)=(0.3,0.5)$} \\ \hline
				\multicolumn{1}{c|}{LID}        &\XSolidBrush              &\XSolidBrush              &\XSolidBrush              & 87.50               & 63.63               & 0.00               \\
				\multicolumn{1}{c|}{Conf.} &\XSolidBrush              &\XSolidBrush              &\XSolidBrush              & 37.50               & 25.00               & 0.00               \\
				\multicolumn{1}{c|}{Loss}       &\XSolidBrush              &\XSolidBrush              &\XSolidBrush              & 78.80               & 60.57               & 0.00               \\
				\multicolumn{1}{c|}{Loss}       & \Checkmark &\XSolidBrush              &\XSolidBrush              & 89.41               & 80.46               & 0.00               \\
				\multicolumn{1}{c|}{Loss}       &\XSolidBrush              & \Checkmark &\XSolidBrush              & 65.77               & \textbf{100.00}     & 37.58              \\
				\multicolumn{1}{c|}{Loss}       & \Checkmark & \Checkmark &\XSolidBrush              & 81.10               & \textbf{100.00}     & 47.09              \\
				\multicolumn{1}{c|}{Loss}       & \Checkmark & \Checkmark & \Checkmark & \textbf{90.23}      & \textbf{100.00}     & \textbf{88.82}     \\ \hline
				\bottomrule
			\end{tabular}
		}
		\caption{Ablation study of the first stage in FedNoRo.}
		\label{tab:ablation1}
	\end{table}
	
	\subsubsection{How Warm-Up Training Affects Noisy Client Detection} \label{t1}
	
\begin{figure}[!t]
		\centering
		\includegraphics[width=1\columnwidth]{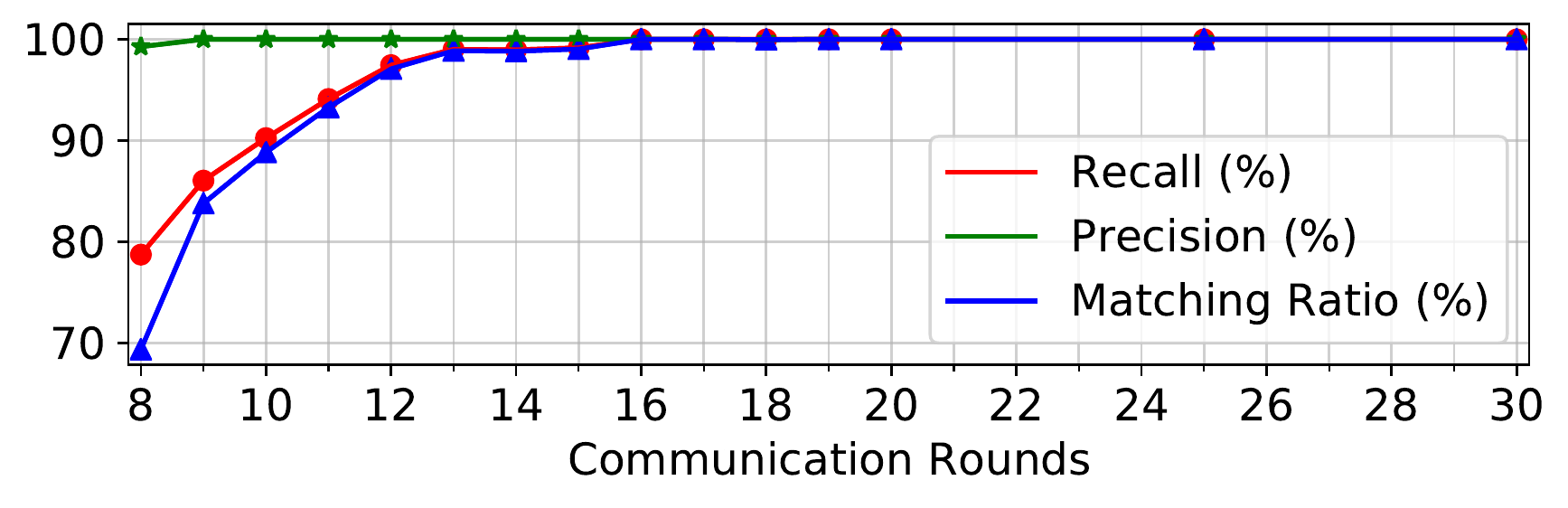}
		\caption{Ablation study of $T_1$ for warm-up training.}
		\label{rounds}
	\end{figure}

	In the first stage, the global model warms up for $T_1$ rounds with FedAvg before noisy client detection. Due to the lack of prior information on noisy clients in realistic scenarios, it is hard to exactly determine $T_1$ for optimal performance. 
	To evaluate the effect of $T_1$, noisy client detection is conducted under different settings of $T_1$  on the ICH dataset ($\rho=0.4$, ($\eta^l,\eta^u$)=(0.3,0.5)) as shown in Fig. \ref{rounds}. 
	It is observed that, after a certain number of warm-up training rounds, noisy client detection performance becomes stable in a certain range. In other words, the setting of $T_1$ would not significantly affect FedNoRo's performance.
	\begin{table}[t]
		\centering
		\renewcommand{\arraystretch}{0.7}
		\resizebox{\columnwidth}{!}{ 
			\begin{tabular}{c|cc|c|c|c}
				\toprule
				\hline
				\multirow{2}{*}{Noisy Clients} & \multicolumn{2}{c|}{De-Noise Strategy} & \multirow{2}{*}{Type} & \multicolumn{2}{c}{BACC (\%)} \\ \cline{2-3}\cline{5-6}
				& \multicolumn{1}{c|}{$\mathcal{L}_{KL}$} & DaAgg            &      & ICH                            & ISIC     \\ \hline
				\multirow{2}{*}{\Checkmark}                                                                    & \multicolumn{1}{c|}{\multirow{2}{*}{\XSolidBrush}}           & \multirow{2}{*}{\XSolidBrush}           & Best                         & 66.60                                                                                                    & 61.20                                                                                                          \\
				& \multicolumn{1}{c|}{}                                       &                                        & Last                         & 63.90                                                                                                    & 56.11                                                                                                          \\ \hline
				\multirow{2}{*}{\XSolidBrush}                                                                  & \multicolumn{1}{c|}{\multirow{2}{*}{\XSolidBrush}}           & \multirow{2}{*}{\XSolidBrush}           & Best                         & 69.67                                                                                                    & 64.94                                                                                                          \\
				& \multicolumn{1}{c|}{}                                       &                                        & Last                         & 68.07                                                                                                    & 59.17                                                                                                          \\ \hline
				\multirow{2}{*}{\Checkmark}                                                                    & \multicolumn{1}{c|}{\multirow{2}{*}{\XSolidBrush}}           & \multirow{2}{*}{\Checkmark}             & Best                         & 69.32                                                                                                    & 65.72                                                                                                          \\
				& \multicolumn{1}{c|}{}                                       &                                        & Last                         & 68.18                                                                                                    & 58.60                                                                                                          \\ \hline
				\multirow{2}{*}{\Checkmark}                                                                    & \multicolumn{1}{c|}{\multirow{2}{*}{\Checkmark}}             & \multirow{2}{*}{\XSolidBrush}           & Best                         & 64.53                                                                                                    & 60.79                                                                                                          \\
				& \multicolumn{1}{c|}{}                                       &                                        & Last                         & 64.01                                                                                                    & 43.57                                                                                                          \\ \hline
				\multirow{2}{*}{\Checkmark}                                                                    & \multicolumn{1}{c|}{\multirow{2}{*}{\Checkmark}}             & \multirow{2}{*}{\Checkmark}             & Best                         & \textbf{70.69}                                                                                           & \textbf{66.00}                                                                                                 \\
				& \multicolumn{1}{c|}{}                                       &                                        & Last                         & \textbf{70.14}                                                                                           & \textbf{60.65}                                                                                                 \\ \hline
				\bottomrule
			\end{tabular}
		}
		\caption{Ablation study of the second stage in FedNoRo. Settings: $\rho$=0.4 and ($\eta^l,\eta^u$)=(0.3,0.5) for the ICH dataset; $\rho$=0.6, ($\eta^l,\eta^u$)=(0.5,0.7) for the ISIC dataset.}
		\label{tab:ablation2}
	\end{table}
	
	\subsubsection{How ${\cal L}_{KL}$ and DaAgg Benefit Noise-Robust Training}
	
	Quantitative results of FedNoRo coupled with various de-noise strategies are summarized in Tab. \ref{tab:ablation2}. 
	Without de-noise training, noisy clients with low-quality local models would surely degrade the classification performance. Though relying only on clean clients can get rid of label noise and somewhat improve classification performance, it may miss valuable information from other clients, making the performance sub-optimal. With noisy clients, replacing FedAvg with DaAgg effectively reduces the weights of noisy clients for global model updating, leading to performance improvement. 
	With noisy clients, only adopting $L_{KL}$ in noisy clients is less effective, as the global model may be severely affected by noisy clients containing inaccurate soft labels. Comparatively, jointly using ${\cal L}_{KL}$ and DaAgg can better explore valuable information from noisy clients while minimizing negative impact, outperforming the classification performance relying only on clean clients.
	
	\section{Conclusion}
	This paper presents a new federated label noise model where global data is class-imbalanced and label noise across clients is heterogeneous. Compared to existing label noise models assuming balanced global data, the proposed noise model is more realistic and flexible to describe real-world FL scenarios. To address this, we further propose a novel two-stage framework named FedNoRo for noise-robust federated learning. Experimental results on publicly-available datasets demonstrate FedNoRo's superior performance against the state-of-the-art FNLL approaches. We believe both the noise model and the solution would inspire future work on developing applicable FL frameworks for real-world applications.
	
	\section*{Acknowledgments}
	This work was supported in part by the National Natural Science Foundation of China under Grants 62202179 and Grant 62271220, in part by the Natural Science Foundation of Hubei Province of China under Grant 2022CFB585, in part by the National Natural Science Foundation of China/Research Grants Council Joint Research Scheme under Grant N\_HKUST627/20, and in part by the Research Grants Council GRF Grant 16203319. The computation is supported by the HPC Platform of HUST.
	
	\bibliographystyle{named}
	\bibliography{ijcai23}
	
	\appendix
	\section*{Appendix}
	
	\section{Setting Comparison}
	Since we conduct experiments on the proposed FNLL problem, our experiments has some differences in setting compared to previous studies (\textit{e.g.} FedCorr \cite{xu2022fedcorr}) as shown in Tab. \ref{tab:setting}.
	
	\begin{table}[h]
		\centering
		\resizebox{\columnwidth}{!}{ 
			\begin{tabular}{lccc}
				\toprule
				\hline
				& Global Class Imbalance & Data Heterogeneity & Label Noise Heterogeneity \\ \hline
				FedCorr & \XSolidBrush                      & \Checkmark                  & \XSolidBrush                         \\
				Ours    & \Checkmark                      & \Checkmark                  & \Checkmark                         \\ \hline
				\bottomrule
			\end{tabular}
		}
		\caption{Comparison with experimental setting from the previous study.}
		\label{tab:setting}
	\end{table}
	
	\section{Data Partition}
	To simulate data heterogeneity, we distribute heterogeneous data to different clients following the previous FNLL study \cite{xu2022fedcorr}. Specifically, a matrix $\Phi \in \{0,1\}^{K \times C}$ is first created, where each element $\Phi_{i,j}$ which is sampled from the element-wise Bernoulli distribution indicates whether client $i$ has samples of class $j$ (1 means yes). Then, for each class $c$, Dirichlet distribution determined by the parameter $\alpha_{Dir}$ is employed to generate a probability distribution to distribute samples of class $c$ to clients that own samples of class $c$.
	
	\section{Implementation Details of Comparative Methods}
	We briefly describe some important implementation details of comparative methods in the experiments.
	\begin{itemize}
		\item FedAvg \cite{mcmahan2017communication}: Vanilla cross entropy loss is employed in the local training phase.
		
		\item FedProx \cite{FedProx}: The proximal term $\mu_{prox}$ is set as 0.01 and 0.1 for ICH and ISIC2019, respectively.
		
		\item FedLA \cite{menon2020long}: Logit adjustment is employed in the local training phase.
		
		\item RoFL \cite{yang2022robust}: $T_{pl}$ is as 10 while other parameter is same as the open source code.
		
		\item RHFL \cite{fang2022robust}: Since the experiment lacks a public dataset for knowledge distillation, client confidence re-weighting which originally designed for knowledge distillation is combined with FedAvg. 
		
		\item FedLSR \cite{jiang2022towards}: $t_w$ is set as 20 and 60 for ICH and ISIC2019, respectively. Other parameter is same as the open source code.
		
		\item FedCorr \cite{xu2022fedcorr}: $T_1,T_2,T_3$ are set as 2, 50, 50 and 5, 150, 150 for ICH and ISIC2019, respectively. One for each round of online clients in the first stage, and all clients online in each of the last two stage.
		
		\item FedCorr+LA: Logit adjustment is employed in the local training phase of FedCorr.
	\end{itemize}
	
	For fair comparasion, all methods share settings not specifically mentioned above for each dataset, \textit{e.g.} backbone model, communication round, local epoch, optimizer (mentioned in Sec. 5.2). Note that to simplify the experimental setup, all clients are online in each round for all methods, except for the first stage of FedCorr due to its special design. To ensure adequate training for FedCorr, the first stage of FedCorr is not counted in the total communication rounds.

\end{document}